\documentclass[10pt,twocolumn,letterpaper]{article}

\usepackage{cvpr}              
\definecolor{cvprblue}{rgb}{0.21,0.49,0.74}
\usepackage[pagebackref,breaklinks,colorlinks,allcolors=cvprblue]{hyperref}

\def\paperID{*****} 
\def\confName{CVPR}
\def\confYear{2026}

\title{FastSHADE: Fast Self-augmented Hierarchical Asymmetric Denoising for Efficient inference on mobile devices}

\author{Nikolay Falaleev\\
Fanis\\
London, UK\\
{\tt\small n.falaleev@fanis.ai}
}

\begin{document}
\maketitle
\begin{abstract}
Real-time image denoising is essential for modern mobile photography but remains challenging due to the strict latency and power constraints of edge devices. This paper presents FastSHADE (Fast Self-augmented Hierarchical Asymmetric Denoising), a lightweight U-Net-style network tailored for real-time, high-fidelity restoration on mobile GPUs. Our method features a multi-stage architecture incorporating a novel Asymmetric Frequency Denoising Block (AFDB) that decouples spatial structure extraction from high-frequency noise suppression to maximize efficiency, and a Spatially Gated Upsampler (SGU) that optimizes high-resolution skip connection fusion. To address generalization, we introduce an efficient Noise Shifting Self-Augmentation strategy that enhances data diversity without inducing domain shifts. Evaluations on the MAI2021 benchmark demonstrate that our scalable model family establishes a highly efficient speed-fidelity trade-off. Our base FastSHADE-M variant maintains real-time latency ($<$50 ms on an Adreno 840 GPU) while preserving structural integrity, and our scaled-up FastSHADE-XL establishes a new state-of-the-art for overall image quality, achieving 37.94 dB PSNR.

\end{abstract}    
\section{Introduction and Related Work}
\label{sec:intro}

While mobile photography has become dominant, it relies heavily on small, high-megapixel camera sensors. This physical limitation results in inherently high levels of noise compared to larger-sensor cameras, such as full-frame DSLRs and mirrorless cameras. As a result, modern mobile photography and videography rely on deep learning for Image Signal Processing (ISP) tasks, with image denoising being among the central components for high-quality capture in suboptimal conditions. However, processing frames directly on mobile edge devices under strict latency and power constraints remains a challenging open problem.

\begin{figure}
  \centering
  \includegraphics[width=\linewidth]{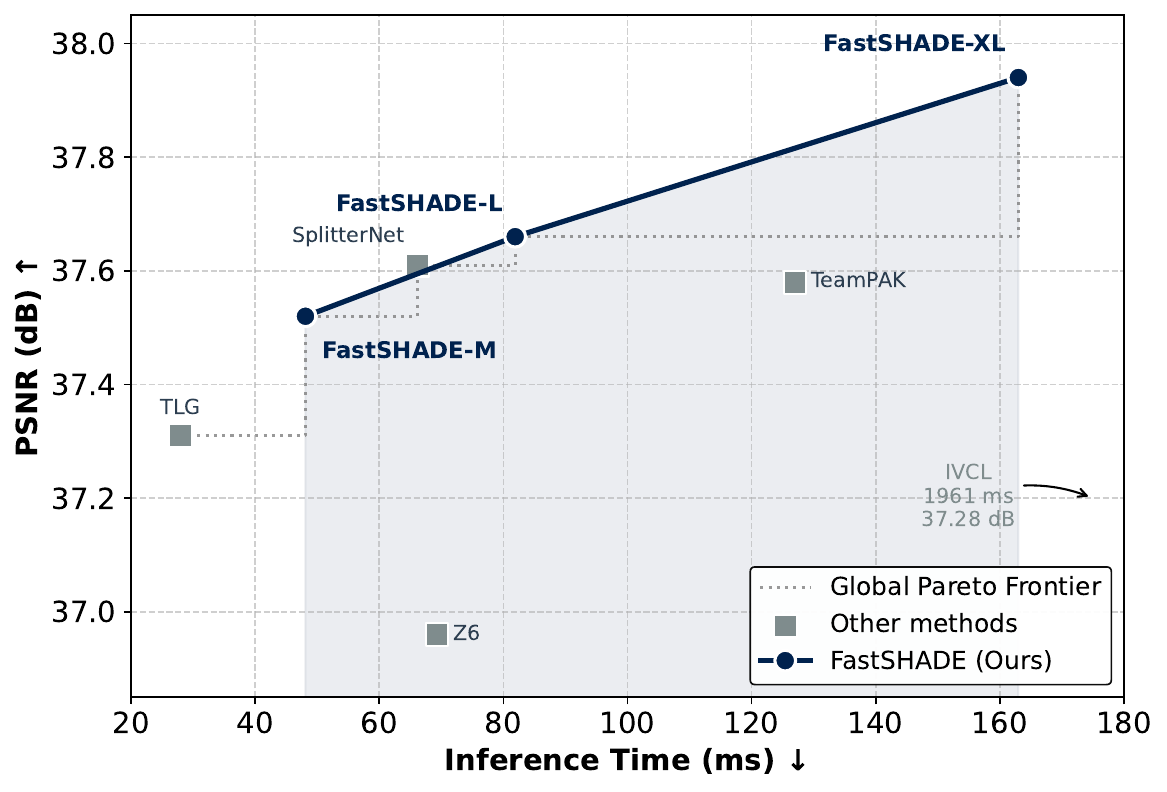}
  \caption{\textbf{Latency-Quality Pareto Frontier.} The FastSHADE model family establishes a highly efficient trade-off between on-device inference time (measured in FP16 on a Qualcomm Adreno 840 GPU) and pixel-level restoration fidelity (PSNR). While ultra-lightweight methods like TLG operate faster, they suffer a noticeable degradation in image quality. In contrast, our scalable architecture significantly expands the state-of-the-art efficiency frontier, pushing the boundaries of visual fidelity while offering highly competitive, low-latency variants for real-world mobile denoising.}
  \label{fig:pareto}
\end{figure}

Historically, classical non-local collaborative filtering algorithms like BM3D~\cite{BM3D_original, BM3D} were applied to this task. It was later shown that learning-based methods are far superior, with standard Convolutional Neural Networks (CNNs) like DnCNN~\cite{DnCNN} and FFDNet~\cite{FFDNet} establishing the baseline for deep image restoration. Early deep learning approaches primarily focused on removing synthetic Additive White Gaussian Noise (AWGN)~\cite{NTIRE2023}. However, as highlighted by recent works~\cite{SplitterNet, SIDD}, synthetic noise models underperform in capturing the true signal-dependent nature of real sensor noise, which is further spatially correlated by the camera's ISP. 

Addressing this domain gap directly on mobile edge devices has become a critical priority for the smartphone industry. This problem's importance is highlighted by prominent open challenges, such as the Mobile AI (MAI)~\cite{MAI2021}, NTIRE~\cite{NTIRE2023}, and AIM workshops hosted at top-tier vision conferences (CVPR, ECCV, ICCV). These competitions explicitly target the Pareto frontier of visual fidelity and actual on-device execution latency, driving the research community away from theoretical FLOPs optimization toward addressing real-world hardware limitations, such as memory bandwidth bottlenecks, limited compute, and restricted energy resources. 

To meet these strict benchmarks, many highly efficient and top-performing models, such as those from Megvii, NOAHTCV, and others~\cite{MAI2021, SplitterNet}, heavily utilize U-Net-style~\cite{UNet} architectures to balance spatial context and computational load. Recent research, for example, NAFNet~\cite{NAFNet}, demonstrated that removing traditional non-linear activation functions can drastically reduce computational complexity, establishing a strong baseline for efficient real-world restoration. Its progressive continuation, CascadedGaze~\cite{CascadedGaze}, further improved this approach by utilizing cascaded bottleneck blocks to handle high-resolution inputs. Another notable approach is SplitterNet~\cite{SplitterNet}, which was optimized for mobile inference by applying feature map channel splitting at different spatial scales to reduce compute. 

Recent advances in state-of-the-art image restoration have also been widely driven by Vision Transformers (ViTs), such as Restormer~\cite{Restormer} and SwinIR~\cite{SwinIR}. While these models achieve superior perceptual metrics, their core mechanism of Self-Attention scales quadratically with image resolution. Even with advanced patch-partitioning schemes, Transformer-based architectures face significant challenges in mobile environments: severe memory access overheads and latency penalties make such approaches prohibitively slow for real-world mobile scenarios.

Consequently, highly efficient CNNs remain the standard for edge deployment. However, we argue that even modern U-Net CNNs waste valuable computational budgets by relying on symmetric convolutions that treat broad, low-frequency structural details and high-frequency, localized noise identically. Furthermore, standard skip connections in U-Net-like architectures inflate channel dimensions, further slowing down mobile inference.

To address these problems, we propose FastSHADE (Fast Self-augmented Hierarchical Asymmetric Denoising), a heavily optimized, edge-deployable denoising architecture. FastSHADE is designed at both the macro-architectural and micro-block levels to maximize representational expressivity during training while providing optimized edge GPU deployment. We achieve this deployment efficiency via structural reparameterization, a technique that decouples training-time topology from deployment-time architecture. RepVGG~\cite{RepVGG} demonstrated that complex multi-branch network blocks can be linearly fused into single convolutions during inference, allowing faster inference without a reduction in prediction accuracy. This concept was effectively adapted for mobile image enhancement in MobileIE~\cite{MobileIE} via the Multi-Branch Re-parameterized Convolution (MBRConv). FastSHADE adopts the MBRConv as its core operator, applying precision modifications to prevent numerical drift during the fusion process. 

In designing FastSHADE, we initially utilized a fixed Haar wavelet transform as a downscaling layer to separate frequencies losslessly. However, we moved beyond fixed transforms, replacing them with learnable $2 \times 2$ strided convolutions that learn a task-specific basis. Furthermore, we introduce an asymmetric block design and a spatially modulated upsampling strategy to handle structural and noise frequencies separately while keeping inference latency low.

In summary, our main contributions are as follows:
\begin{itemize}
\item We propose FastSHADE, an efficient, structural-reparameterization-based U-Net architecture tailored for mobile GPUs. Our scalable model family effectively expands the speed-fidelity Pareto frontier for real-time edge denoising.
\item We introduce the Asymmetric Frequency Denoising Block (AFDB), a novel micro-architecture that optimizes computational resource allocation by decoupling heavy spatial structure extraction from rapid, high-frequency noise suppression.
\item We design the Spatially Gated Upsampler (SGU), an efficient upsampling mechanism that dynamically modulates high-resolution skip connections to improve fine-grained structural recovery.
\item We present a Noise Shifting Self-Augmentation strategy that improves data diversity without inducing domain shifts. Combined with our architecture, FastSHADE-M achieves a highly practical balance of fidelity and latency, while FastSHADE-XL establishes a new state-of-the-art for overall image quality on the MAI2021 dataset.
\end{itemize}
\section{Proposed Method}
\label{sec:method}

\begin{figure*}
  \centering
   \includegraphics[width=0.90\linewidth]{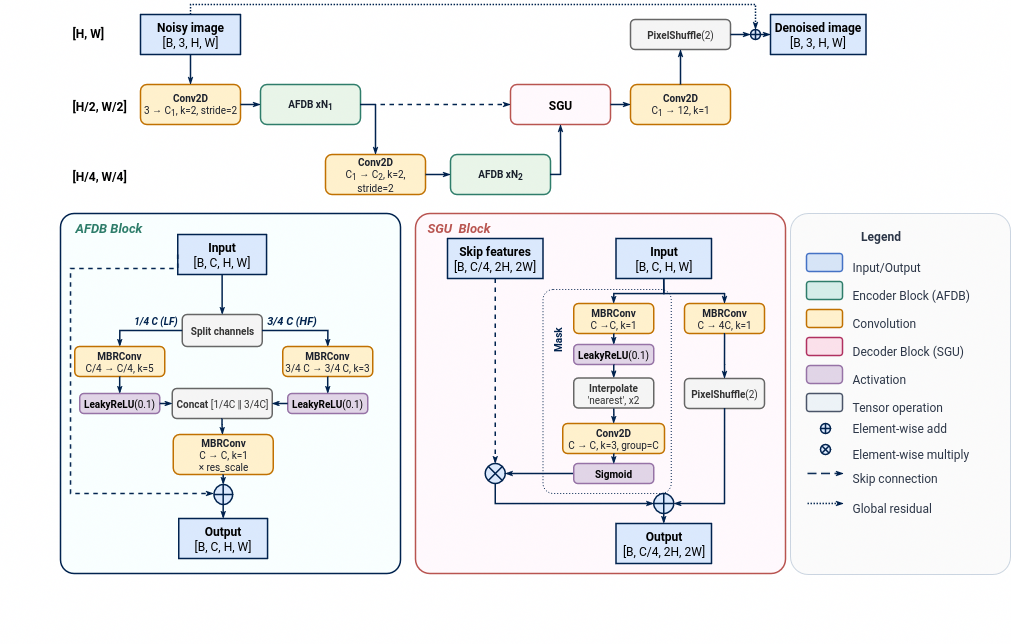}
  \caption{Proposed model architecture}
  \label{fig:architecture}
\end{figure*}

We propose FastSHADE, a U-Net-style~\cite{UNet} model, depicted in \cref{fig:architecture}. Given an input noisy image $\mathbf{I}_{noisy} \in \mathbb{R}^{3 \times H \times W}$, the network aims to predict the clean residual $\mathbf{R} \in \mathbb{R}^{3 \times H \times W}$ such that the output $\mathbf{I}_{clean} = \mathbf{I}_{noisy} + \mathbf{R}$. The model internally accounts for data normalization, allowing it to directly ingest and output RGB images with integer pixel values in the range of $[0, 255]$, while the model operates in FP16 precision during inference.

\subsection{Architecture}
Originally, the model used a fixed Haar Wavelet Transform for spatial downsampling; however, based on experimental results, FastSHADE now replaces it with learnable $2 \times 2$ convolutions with a stride of 2. The network computes representations at two spatial scales, projecting features to channels $C_1$ and $C_2$. 

The downsampled representations are processed using a sequence of Asymmetric Frequency Denoising Blocks (AFDB). We use $N_1$ intermediate blocks at the first scale and $N_2$ blocks at the lowest spatial resolution. Finally, features are progressively restored to the original resolution using a Spatially Gated Upsampler (SGU) and standard PixelShuffle operations.

Parameterizing the architecture with $C_1$, $C_2$, $N_1$, and $N_2$ allows for the creation of various model sizes, which balance latency and visual quality of denoised images.

\subsection{Asymmetric Frequency Denoising Block (AFDB)}

Natural image noise is highly frequency-dependent. We argue that standard convolutions often waste computation by processing low-frequency structural details and high-frequency noise representations identically. To address this, we propose the Asymmetric Frequency Denoising Block (AFDB), which, unlike the channel bifurcation approach of SplitterNet~\cite{SplitterNet}, splits the incoming feature maps into two asymmetrical paths along the channel dimension:

\begin{enumerate}
    \item \textbf{Low-Frequency (LF) Spatial Path:} $25\%$ of the channels are routed to a branch parameterized by a large-kernel convolution ($5 \times 5$). This branch acts as a heavy spatial feature extractor, responsible for recovering broader structural contexts and edges.
    \item \textbf{High-Frequency (HF) Noise Path:} The remaining $75\%$ of the channels are routed to a branch with a smaller kernel size convolution ($3\times3$). This branch is designed for rapid noise suppression.
\end{enumerate}

Both branches use a LeakyReLU ($\alpha=0.1$). Following feature extraction, features from both branches are concatenated and fused via a $1 \times 1$ convolution. To stabilize the training dynamics and facilitate gradient flow, we apply a learnable residual scaling parameter (initialized to $0.2$) and add the input feature map via a skip connection.

\subsection{Spatially Gated Upsampler (SGU)}

To mitigate the artifacts often induced by spatially upscaling layers, such as transposed convolutions, we utilize PixelShuffle for upscaling with added high-resolution skip features. However, naively concatenating high-resolution skip features with upsampled low-resolution features can degrade mobile efficiency. Instead, we introduce a Spatially Gated Upsampler (SGU) that dynamically modulates the skip connection, allowing the model to selectively propagate features from different scales. 

Given a low-resolution feature $\mathbf{F}_{LR}$ and a high-resolution skip feature $\mathbf{F}_{HR}$, the SGU generates a spatial modulation mask:
\begin{equation}
    \mathbf{M} = \sigma\Big( \text{Conv}_{3\times3}^{dw}\big(\text{Up}_{2\times}(\text{LReLU}(\text{Conv}_{1\times1}(\mathbf{F}_{LR})))\big) \Big)
\end{equation}
where $\sigma$ is the sigmoid function, $\text{Conv}_{3\times3}^{dw}$ is a depthwise convolution for the mask refinement, and $\text{Up}_{2\times}$ denotes 2x nearest-neighbor interpolation. The upsampled feature is then combined as:
\begin{equation}
    \mathbf{F}_{out} = \text{PixelShuffle}_2(\text{Conv}_{1\times1}(\mathbf{F}_{LR})) + \mathbf{F}_{HR} \odot \mathbf{M}
\end{equation}
This multiplicative gating effectively acts as a spatial attention mechanism, only allowing necessary structural details to propagate. This boosts the PSNR without increasing the channel dimension, which is beneficial for faster mobile GPU inference.

\subsection{Noise Shifting Self-Augmentation}

Standard noise augmentation strategies often inject synthetic noise, which can introduce a severe domain gap by pushing the training data off the true noise distribution. Drawing inspiration from Recorrupted-to-Recorrupted (R2R)~\cite{R2R} and SRDTrans~\cite{SRDTrans} training approaches, we propose Noise Shifting Self-Augmentation. Instead of synthesizing artificial noise distributions, we generate a statistically valid alternative noise sample directly from the dataset images. We deliberately avoid swapping noise profiles from other images, even those captured by the same sensor, as real-world sensor noise is highly sensitive to specific camera settings, lighting conditions, and the scene content.

Given a clean ground-truth image $\mathbf{x}_{\text{clean}}$ and its noisy observation $\mathbf{x}_{\text{noisy}}$, we first extract the exact true sensor noise residual $\mathbf{N} = \mathbf{x}_{\text{noisy}} - \mathbf{x}_{\text{clean}}$. Assuming that natural images are locally smooth, the signal-dependent noise variance (such as Poisson-Gaussian shot noise) and ISP spatial correlations remain statistically stable within a radius of several pixels. Using this property, we create a valid alternate noise sample $\mathbf{N}_{alt}$ by applying a small random spatial translation ($s_x, s_y$ pixels) with reflection padding to the original noise:
\begin{equation}
    \mathbf{N}_{alt} = \text{Shift}(\mathbf{N}, s_x, s_y)
\end{equation}
where the spatial shifts are bounded such that $s_x, s_y \in [-2, 2] \setminus \{0\}$. 

Because $\mathbf{N}_{alt}$ is a pure spatial translation of the true noise, it inherently preserves the overall statistics and the realistic covariance structure of the original noise signature. This eliminates the need for complex spatial error masking or noise level profiling. The final augmented input is directly reconstructed by applying this shifted noise sample to the ground truth:
\begin{equation}
\label{eq:noise_shift}
    \tilde{\mathbf{x}}_{\text{noisy}} = \text{Clip}(\mathbf{x}_{\text{clean}} + \mathbf{N}_{alt}, \, 0, \, 255)
\end{equation}
This simple yet effective self-augmentation strategy provides new in-distribution noise samples, improves structural generalization, and avoids the out-of-distribution domain shifts associated with synthetic data.

\subsection{Deployment Optimization}

\textbf{Multi-Branch Re-parameterized Convolution:} \textit{MBRConv}, adapted from MobileIE~\cite{MobileIE}, is used as the foundational convolutional layer in the model's blocks. During training, these blocks operate as heavy Multi-Branch Re-parameterized Convolutions (MBRConv) with an expansion ratio of $\gamma=4$. For inference on mobile devices, the multi-branch topologies are re-parameterized into a single dense convolution during the export phase, yielding computationally equivalent results. Unlike the original MobileIE, we utilize double-precision (FP64) float conversion during this fusion step to minimize numerical errors in the weight calculations.

\textbf{Quantization and Folding:} Although the model is inferred using FP32 weights with FP16 computations, the objective is to ingest and predict integer pixel values directly in the range $[0, 255]$ without rounding-related errors, thereby avoiding extra computational overhead when processing real-world 8-bit images. To achieve this, we employ a Straight-Through Estimator (STE) as defined by \cref{eq:STE} during training for gradient-preserving rounding quantization of the model's output tensor.

\begin{equation}
\mathbf{I}_{out} = \mathbf{I}_{out} + (\text{Round}(\mathbf{I}_{out}) - \mathbf{I}_{out})\text{.detach()}
\label{eq:STE}
\end{equation}

Furthermore, the model operates on data normalized to $[0, 1.0]$ range to facilitate normal distribution of Batch Normalization operations, embedded in MBRConv Blocks. During deployment export, the input normalization scalar ($1/255.0$), output denormalization scalar ($\times 255.0$) and the learnable residual scaling are analytically fused into the weights of the corresponding convolutional layers to avoid these operations overhead during inference.

\section{Experimental setup}
\label{sec:experiments}
\begin{figure*}
  \centering
   \includegraphics[width=0.95\linewidth]{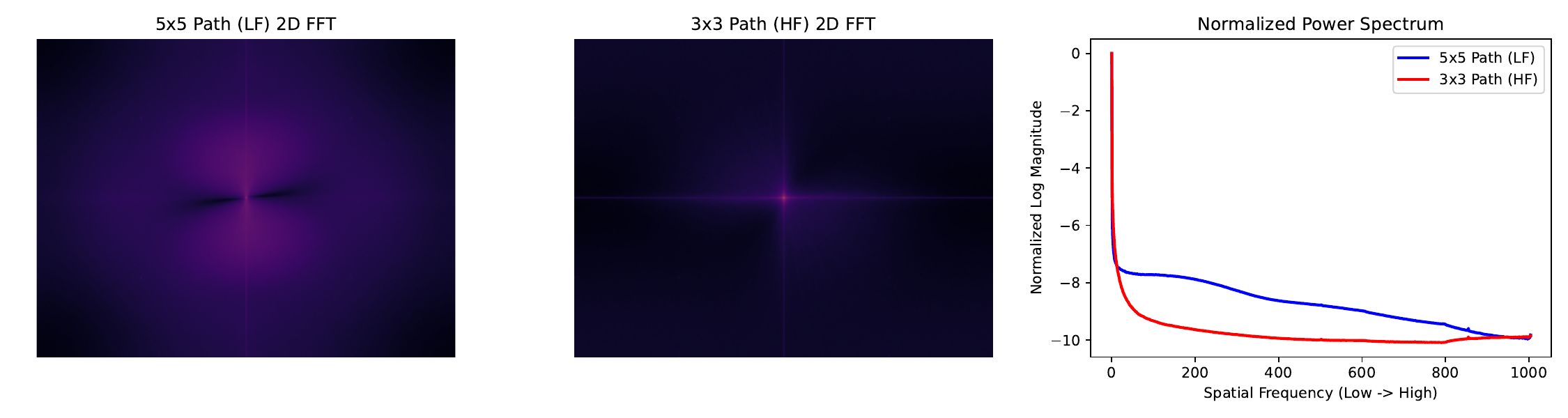}
  \caption{Frequency domain analysis of the internal features within the AFDB Block, averaged over 50 validation images. (Left, Center) The 2D FFT magnitude spectra exhibit a broader energy distribution in the $5\times5$ path compared to the $3\times3$ path. (Right) In the normalized radially averaged power spectrum, the spatial path (blue) retains significant energy in the low-to-mid spatial frequencies, indicating the preservation of structural details and edges. In contrast, the HF noise path (red) has a flat spectral response characteristic of isotropic, unstructured noise extraction.}
  \label{fig:AFDB_spectrum}
\end{figure*}

\subsection{Implementation details}

Model training was performed using the PyTorch 2.9 framework. The models were converted into TFLite format for mobile device runtime evaluation using litert-torch v0.8.0 \cite{litert}. The model format conversion also included tensor axis rearrangement to allow channel-last memory layout. The final model is exported in the FP32 format and evaluated with the GPU delegate on the mobile device in FP16 computation mode.

\subsubsection{Model training protocol}
The model was trained using the AdamW optimizer~\cite{AdamW} with \( \beta_1 = 0.9, \beta_2 = 0.999 \) with a Cosine Annealing learning rate schedule decaying from $1 \times 10^{-3}$ to $1 \times 10^{-5}$. The training procedure consisted of two stages:
\begin{itemize}
    \item Model training for 80 epochs and Charbonnier loss (\cref{eq:Charbonnier}, where $\hat{x}$ is the predicted denoised image, $y$ is the ground truth image and $\epsilon=10^{-3}$) on a batch size of 32 samples, each consisting of $256 \times 256$ pixel patches.
    \item Fine-tuning for 20 epochs with PSNR loss (\cref{eq:PSNR_loss}, where $max_y$ is the maximal possible value of $y$, which is 255) using $512 \times 512$ px patches. The loss function was used as it directly optimizes the target metric. We observed that simpler L1 or MSE losses do not allow achieving the same result.
\end{itemize}

\begin{equation}
\mathcal{L}_{Charbonnier} = \sqrt{(\hat{x} - y )^2 + \epsilon^2},
\label{eq:Charbonnier}
\end{equation}

\begin{equation}
\mathcal{L}_{PSNR}(\hat{x}, y)=-10log\frac{(max_y)^2}{MSE(\hat{x}, y)},
\label{eq:PSNR_loss}
\end{equation}

During training we tracked an exponential moving average (EMA) of the model parameters for both stages, using a decay factor of 0.999, which stabilizes training by smoothing short-term fluctuations in stochastic updates. Final evaluations were performed on the EMA-smoothed weights.

The training data was augmented by random 90-degree rotations with 50\% probability, random horizontal and vertical flipping with 30\% probability. Furthermore, for 20\% of the samples, the input image was replaced with the ground-truth image artificially corrupted by random Gaussian noise with $\sigma=25$. Additionally, we utilized our self-augmentation strategy with probability of 25\%.

\subsection{Datasets}

The main dataset utilized in the experiments was MAI2021~\cite{MAI2021}, employing its training subset (658 images of $3000\times4000$ pixel resolution) for models training, and its validation part (50 images) for fidelity evaluation. Ground truth denoised images of the dataset were created via burst mode: for each scene, 20 images were captured and averaged to get a clean corresponding photo.

Additionally, training parts of SIDD-Medium~\cite{SIDD} and MIDD~\cite{SplitterNet} datasets were used for training. Training was performed on patches extracted from the original training images using a sliding window cropping.

The MIDD dataset contains data captured using 20 different sensors, so we constructed batches as follows: of the 32 patches per batch, 10 were from the MAI2021 dataset, followed by 20 patches from MIDD (ensuring 1 patch from each sensor type), and finally 2 patches from the SIDD dataset.

\subsection{Evaluation}

SSIM and PSNR metrics were used as the main fidelity metrics. Additionally, following \cite{MAI2021}, we adopted the score \cref{eq:score}, which balances inference time and PSNR value.

\begin{equation}
s=\frac{2^{2(PSNR-38.0)}}{t},
\label{eq:score}
\end{equation}
where $t$ is inference time, obtained on a mobile device.

In our evaluations, we used mobile devices with ARM Mali-G78 MP20 and Qualcomm Adreno 840 mobile GPUs. All mobile runtime results are reported for images in $1088\times1920$ px resolution using AI Benchmark v6.0.5 \cite{ai-benchmark}.

\section{Results and discussion}

\subsection{Model results}

To explore the Pareto frontier of denoising fidelity and on-device latency, we propose three scalable variants of our architecture: \textbf{FastSHADE-M} (base model: $C_1=16$, $C_2=64$, $N_{1}=2$, $N_{2}=3$), \textbf{FastSHADE-L} ($C_1=24$, $C_2=96$, $N_{1}=2$, $N_{2}=3$), and \textbf{FastSHADE-XL} ($C_1=32$, $C_2=128$, $N_{1}=3$, $N_{2}=5$).

\begin{figure*}
  \centering
   \includegraphics[width=0.87\linewidth]{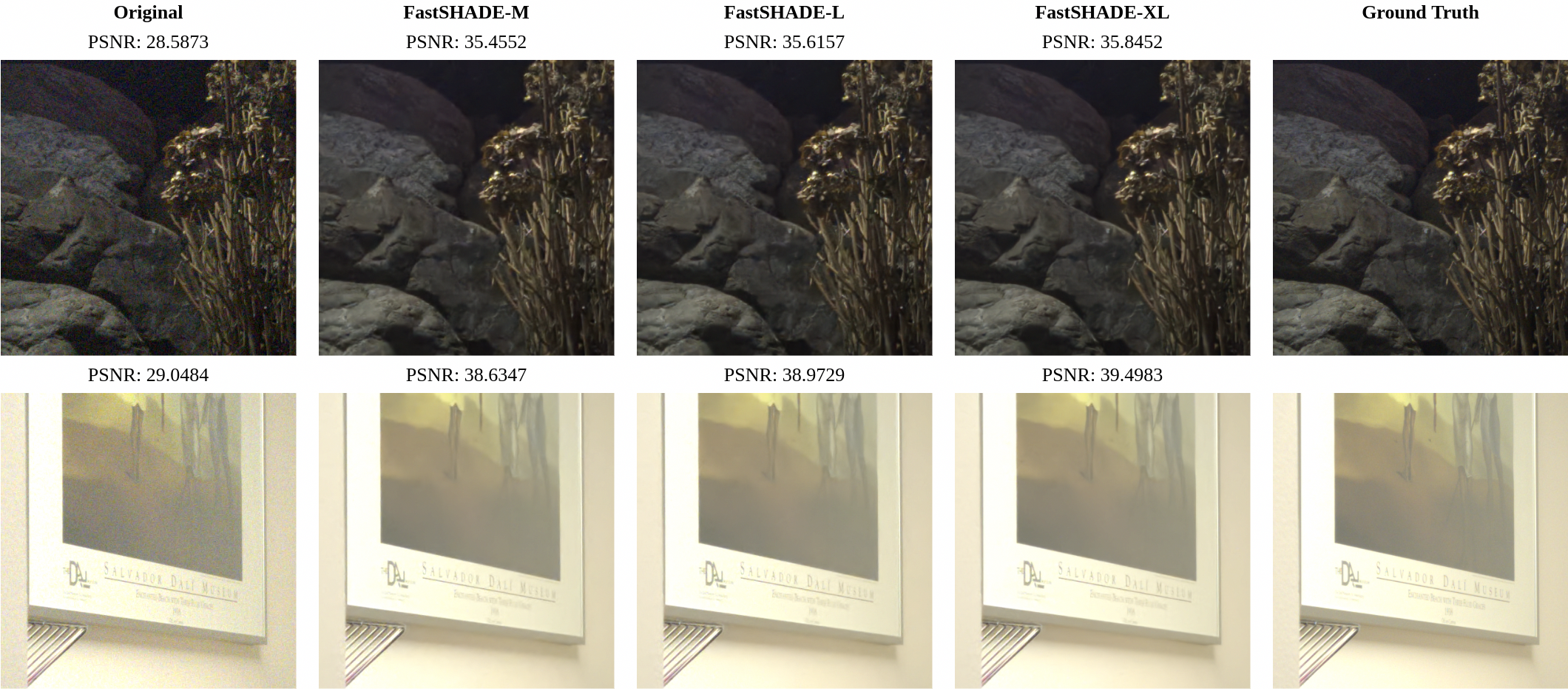}
  \caption{Qualitative results of FastSHADE model predictions.}
  \label{fig:results}
\end{figure*}

\Cref{tab:model_results} and \cref{fig:pareto} summarize the performance of FastSHADE against baselines and top-performing submissions from the Mobile AI 2026 Real Image Denoising Challenge (evaluated on the MAI2021 test subset), demonstrating the efficiency of our approach. On the target Adreno 840 mobile GPU, our base model, FastSHADE-M, achieves an inference time of just 48.1 ms. This makes it significantly faster than competitive baselines like SplitterNet~\cite{SplitterNet} (66.1 ms) and TeamPAK (127 ms), while maintaining highly competitive denoising performance (PSNR of 37.52 dB on the test set). While the ultra-lightweight method TLG achieves the highest challenge score by minimizing latency, this comes at the cost of noticeable denoising quality degradation (37.31 dB). FastSHADE was designed for real-world smartphone pipelines, where perceptual quality cannot be severely compromised. Consequently, FastSHADE-M provides a much more practical balance, yielding a competitive challenge score of 10.68 on the target hardware.

Furthermore, our scaled-up variants demonstrate the significant representational capacity of the AFDB. When latency constraints are relaxed, FastSHADE-\{L, XL\} demonstrate state-of-the-art image denoising results, with FastSHADE-XL achieving the highest overall PSNR and SSIM among all compared methods. Prediction results for the FastSHADE model family are presented in \cref{fig:results}.

\begin{table}
  \centering
  \caption{Performance metrics of FastSHADE compared to baselines and other methods. SSIM and PSNR values are reported for the MAI2021 test subset. Runtime, $t$, is reported for the Adreno 840. SSIM for SplitterNet was not reported because the test subset is not public.}
  \begin{tabular}{lcccc}
    \toprule
    \textbf{Method} & \textbf{SSIM}$\uparrow$ & \textbf{PSNR}$\uparrow$ & $\mathbf{t}$ \textbf{(ms)}$\downarrow$ & \textbf{Score}$\uparrow$ \\
    \midrule
    TLG & 0.9129 & 37.31 & \textbf{28.0} & \textbf{13.72}  \\
    SplitterNet \cite{SplitterNet} & - & 37.62 & 66.1 & 8.93 \\
    TeamPAK & 0.9098 & 37.58 & 127.0 & 4.40  \\
    Z6 & 0.9001 & 36.96 & 69.3 & 3.41  \\
    IVCL & 0.9110 & 37.28 & 1961 & 0.19 \\
    \midrule
    FastSHADE-M & 0.9136 & 37.52 & 48.1 & 10.68 \\
    FastSHADE-L & 0.9150 & 37.66 & 81.9 & 7.61 \\
    FastSHADE-XL & \textbf{0.9179} & \textbf{37.94} & 163 & 5.64 \\
    \bottomrule
  \end{tabular}
  \label{tab:model_results}
\end{table}

\subsection{AFDB features}

While no explicit frequency constraints are applied in the AFDB or loss function, to validate the implicit empirical frequency-routing mechanism of the AFDB, we analyzed the spectral properties of its internal representations. \Cref{fig:AFDB_spectrum} presents the 2D Fast Fourier Transform (FFT) magnitudes and the normalized radially averaged power spectrum of the feature maps from both branches of a trained model, averaged over the validation set. The analysis demonstrates functional differences, which can be explained by architectural inductive biases, without the need for explicit frequency-domain loss functions.

The $5\times5$ branch maintains a significant concentration of energy in the low-to-mid spatial frequencies, visible as a prominent "shoulder" in the power spectrum with a monotonic drop towards higher frequencies. This indicates that the larger receptive field naturally biases the branch toward extracting broad spatial structures, edges, and contextual textures. Conversely, the $3\times3$ branch exhibits a drop in structural frequencies, flattening out almost immediately into a uniform energy distribution. This flat spectral response indicates isotropic, unstructured noise. Thus, the network learns a division of labor: the LF path acts as a heavy spatial feature extractor to preserve image topology, while the highly-parameterized LL path, which receives 75\% of the input channels, is dedicated more to suppressing unstructured noise. This empirical evidence justifies both the frequency-based name of the block and the asymmetric channel design of the proposed AFDB.

\subsection{Ablation Studies}

This section studies the impact of individual components of the proposed approach. To manage computational constraints, all ablation models were trained on a 25\% subset (using every 4th patch) of the MAI2021 training dataset for 40 epochs following the main-stage training protocol. Runtime was evaluated on an ARM Mali-G78 MP20 GPU. We used the FastSHADE-M architecture for these experiments.

\textbf{Downscaling Strategy.} We compare different spatial downscaling strategies in \cref{tab:abl_down}. Originally, our architecture used a discrete Haar wavelet transform to downscale spatial dimensions to losslessly expand the channel capacity. This was implemented as a fixed-weight $2\times2$ convolution with a stride of 2, initialized with Haar basis functions, and followed by a trainable $1\times1$ MBRConv projection layer. However, empirical observations revealed that unfreezing these Haar weights improves the PSNR metric. Building on this conclusion, we replaced the entire two-step Haar-projection block with a single trainable strided convolution that directly maps to the target channel dimension. This modification substantially boosted PSNR (+0.09 dB) and reduced inference time. Finally, we observed that reducing its kernel size from $3\times3$ to $2\times2$ yields virtually identical denoising fidelity but further decreases mobile inference latency, maximizing the target score. Consequently, the $2\times2$ strided convolution was adopted as the optimal downscaling mechanism for the final architecture.

\begin{table}
  \centering
  \caption{Ablation study on spatial downscaling strategies. Replacing the Haar wavelet transform and projection layer with a single learnable $2\times2$ strided convolution achieves the highest score.}
  \label{tab:abl_down}
  \begin{tabular}{@{}lccc@{}}
    \toprule
    \textbf{Downscaling Layer} & \textbf{PSNR(dB)}$\uparrow$ & $\mathbf{t}$\textbf{(ms)}$\downarrow$ & \textbf{Score}$\uparrow$ \\
    \midrule
    Haar (frozen) & 37.475 & 151 & 3.198 \\
    Haar (learnable) & 37.492 & 151 & 3.275 \\
    $3\times3$ Conv (stride 2) & \textbf{37.581} & 141 & 3.967 \\
    \textbf{$2\times2$ Conv (stride 2)} & 37.580 & \textbf{136} & \textbf{4.108} \\
    \bottomrule
  \end{tabular}
\end{table}

\textbf{AFDB Channel Split.} A key feature of the proposed Asymmetric Frequency Denoising Block is its uneven split of channels between branches. We hypothesize that for image denoising, extracting low-frequency (LF) spatial structures requires significantly fewer parameters than filtering high-frequency (HF) noise. To validate this optimized compute resource utilization, we evaluate different channel routing ratios in \cref{tab:abl_split}. A baseline that does not use splitting and routes all features through the heavy $5\times5$ MBRConv achieves suboptimal PSNR and higher latency. Introducing an equal 1:1 channel split between the LF and HF branches slightly improves visual fidelity. However, our proposed asymmetric 1:3 split (allocating only 25\% of channels to the LF branch and 75\% to the HF branch) achieves the exact same PSNR while running faster. This empirical evidence validates our hypotheses, demonstrating the computational redundancy of symmetric processing and justifying the asymmetric design of the AFDB.

\begin{table}[h]
  \caption{Evaluation of channel splitting strategies in the AFDB. 1(LF):0(HF) routes all features through the $5\times5$ spatial branch; 1(LF):1(HF) represents a symmetric split; and 1(LF):3(HF) is our proposed asymmetric configuration (25\% $5\times5$ spatial branch, 75\% $3\times3$ noise). The asymmetric split matches the PSNR of the symmetric baseline but reduces inference time.}
  \label{tab:abl_split}
  \centering
  \begin{tabular}{@{}lccc@{}}
    \toprule
    \textbf{Channel Split LF:HF} & \textbf{PSNR(dB)} $\uparrow$ & $\mathbf{t}$\textbf{(ms)}$\downarrow$ & \textbf{Score}$\uparrow$ \\
    \midrule
    1:0 (No Split) & 36.56 & 144 & 3.773 \\
    1:1 (Symmetric) & \textbf{37.58} & 142 & 3.934 \\
    \textbf{1:3 (AFDB)} & \textbf{37.58} & \textbf{136} & \textbf{4.108} \\
    \bottomrule
  \end{tabular}
\end{table}

\textbf{Spatially Gated Upsampler (SGU) Architecture.} In this ablation (\cref{tab:abl_sgu}), we study architectures employing various attention mechanisms to modulate the high-resolution skip features prior to addition.

As an attention-free baseline, we used standard U-Net concatenation followed by a $1\times1$ convolution. This approach doubles the channel count processed by the following convolution, which adds memory bandwidth penalty during inference, making it the slowest variant. Next, we evaluated global and spatial feature weighting by applying Squeeze-and-Excitation (SE)~\cite{SE} (channel attention) and CBAM-style~\cite{CBAM} (spatial attention) blocks directly to the skip connection. While these methods successfully bypass the channel-doubling bottleneck, roughly halving the inference time compared to the baseline, they yield comparable PSNR. This indicates that channel-level or coarse spatial attention is insufficient to provide the fine-grained guidance for local feature restoration. 

To evaluate dense, per-element attention, we employed a $2\times2$ transposed convolution (with a stride of 2) followed by a sigmoid activation to generate a high-resolution attention mask from the low-resolution latent space. This approach results in a significant +0.14 dB PSNR improvement over the baseline while maintaining low latency, demonstrating that dense per-element gating is crucial for accurately blending features from different scales. 

Finally, we assessed the proposed SGU block with and without its final mask-refinement depthwise convolution ($\text{Conv}_{3\times3}^{dw}$). The results demonstrate that the full SGU block achieves the highest PSNR and the highest overall metric score. Notably, the addition of the depthwise refinement layer improves the final image quality by smoothing the upsampled gating mask, while it does not add measurable latency overhead on the mobile GPU, justifying our final architectural design.

\begin{table}
  \caption{Ablation study on skip connection fusion strategies in the upsampling block: compare the proposed SGU against standard U-Net concatenation, channel/spatial attention (SE and CBAM), and a transposed convolution baseline.}
  \centering
  \begin{tabular}{@{}lccc@{}}
    \toprule
    \textbf{Upsampler Variant} & \textbf{PSNR(dB)} $\uparrow$ & $\mathbf{t}$\textbf{(ms)}$\downarrow$ & \textbf{Score}$\uparrow$ \\
    \midrule
    Concat + $1\times1$ Conv & 37.40 & 257 & 1.690 \\
    SE Attn. & 37.37 & 145 & 2.863 \\
    CBAM Attn.  & 37.43 & 141 & 3.205 \\
    Tconv $2\times2$ + Sigmoid  & 37.54 & 142 & 3.700 \\
    \midrule
    SGU (w/o $\text{Conv}_{3\times3}^{dw}$)  & 37.55 & \textbf{136} & 3.919 \\
    \textbf{SGU}  & \textbf{37.58} & \textbf{136} & \textbf{4.108} \\
    \bottomrule
  \end{tabular}
  \label{tab:abl_sgu}
\end{table}

\textbf{Data Augmentations}. Expanding the training dataset while keeping the artificial examples realistic is crucial for achieving robust results with limited input data. \Cref{tab:abl_data} demonstrates that the addition of basic augmentations substantially improves the PSNR metric, which indicates that the dataset can benefit from data augmentation. The further increase in PSNR demonstrates that our proposed Noise Shifting Self-Augmentation is an efficient addition to traditional image augmentations and is superior to a random swap of noise profiles.

\begin{table}
  \caption{Data augmentation techniques ablation results. \textit{Baseline} means no augmentations; \textit{+ basic} adds random rotations, flips, and artificial Gaussian noise; ($\sigma=25$); \textit{+ basic + noise swap} includes basic augmentation and random noise profile swap between images in a batch; \textit{+ basic + our} adds the proposed Noise Shifting Self-Augmentation.}
  \centering
  \begin{tabular}{@{}lcc@{}}
    \toprule
    \textbf{Data augmentation} & \textbf{PSNR(dB)}$\uparrow$  \\
    \midrule
    Baseline & 36.80  \\
     + basic  & 37.58  \\
     + basic + noise swap & 37.51  \\
     \textbf{+ basic + our}  & \textbf{37.63}  \\
    \bottomrule
  \end{tabular}
  \label{tab:abl_data}
\end{table}
\section{Limitations and Broader Impacts}
\label{sec:limitations}
While FastSHADE demonstrates exceptional Pareto-optimality on mobile GPUs, the current architecture is fundamentally optimized for FP32 and FP16 execution. Quantizing the model to INT8 could result in further speedups and memory reductions; however, this requires quantization-specific training optimizations, such as Quantization-Aware Training, which was not explored in this work. Furthermore, while our model runs efficiently via the standard TFLite GPU delegate, we have yet to exploit low-level hardware optimizations. Future work will focus on tailoring the architecture for specific hardware streams and advanced memory hierarchies, such as explicitly maximizing SRAM utilization and optimizing execution graphs for dedicated Neural Processing Units (NPUs) to fully utilize the available hardware and mitigate memory bandwidth bottlenecks.

The broader impacts of FastSHADE extend beyond image denoising. The core components of our architecture, particularly the AFDB and the SGU block, offer a modular design that is transferable to other mobile vision tasks requiring high fidelity and low latency. For instance, the frequency-aware routing in AFDB is inherently well-suited for \textit{image enhancement} and \textit{low-light enhancement}. Similarly, the SGU mechanism, which efficiently refines upsampled features using gating, can be adapted for other U-Net-style models.

Advancing on-device computational photography offers significant societal benefits. By enabling real-time, high-fidelity image denoising locally, FastSHADE eliminates the need for cloud-based processing, enhancing privacy and security as sensitive photographs remain on-device. Additionally, the model's efficient inference reduces energy consumption.

\section{Conclusion}
\label{sec:conclusion}

We presented FastSHADE, an efficient U-Net-style architecture optimized for real-time image denoising on mobile devices. At the core of our approach is the Asymmetric Frequency Denoising Block, which challenges conventional symmetric channel allocation by routing low-frequency structures and high-frequency noise through tailored processing paths. Combined with our Spatially Gated Upsampler and the proposed Noise Shifting Self-Augmentation strategy, FastSHADE maximizes representational capacity under strict latency constraints. Extensive empirical evaluations demonstrate that our scalable model family effectively expands the efficiency frontier for mobile denoising, enabling fast inference on real-world mobile devices. Our base variant, FastSHADE-M, delivers an optimal balance for real-world deployment, operating noticeably faster than competitive baselines while maintaining exceptional image fidelity. Furthermore, when latency constraints are relaxed, FastSHADE-XL establishes a new state-of-the-art on the MAI2021 dataset, achieving the highest overall PSNR.

\section*{Acknowledgments}

This research was fully self-funded.
We extend our sincere gratitude to Andrey Ignatov of ETH Zürich and the organizers of the Mobile AI 2026 Real Image Denoising Challenge for providing the benchmarking framework and facilitating the evaluations on the target Adreno 840 mobile GPU.

{
    \small
    \bibliographystyle{ieeenat_fullname}
    \bibliography{main}
}


\end{document}